# Face Synthesis (FASY) System for Generation of a Face Image from Human Description

Santanu Halder

Department of Computer Science and Engineering, GNIT
Kolkata - 700114, India
Email: sant.halder@gmail.com

Debotosh Bhattacharjee, Mita Nasipuri,
Dipak Kumar Basu*, Mahantapas Kundu
Department of Computer Science and Engineering,
Jadavpur University, Kolkata, 700032, India
*AICTE Emeritus Fellow

*Abstract*-This paper aims at generating a new face based on the human like description using a new concept. The FASY (FAce SYnthesis) System is a Face Database Retrieval and new Face generation System that is under development. One of its main features is the generation of the requested face when it is not found in the existing database, which allows a continuous growing of the database also.

*Index Terms*- Adjusting Intensities, Assembling the Components, Face Generation, Databases

## I. INTRODUCTION

Face is most important visual identity of a person and while meeting an unknown person, it is the face that attracts our attention most. We often describe a person in terms of the characteristic features of important face components like eyes, eyebrows, nose, lips together with the overall shape of the face. The criminal Investigation agencies employ artists to sketch human faces from the description given by a witness about criminal's appearance which is then searched in the face database of the known criminals for recognition. Since early 1990's, Face Recognition Technology (FRT) becomes an active research area and there are a lot of works on facial recognition and facial feature extraction [1][2][3][4][5][6]. In 1994, Jian Kang Wu and Arcot Desai Narasimhalu [7] worked on "Identifying Faces Using Multiple Retrievals" where they retrieved the existing faces based on some properties of the different face components. But if the desired face is not in the database, then it is necessary to construct a new face from the user's description and this work is a new one which is one of the objective of the present work. This paper is a part of a main research effort, whose aim is the construction of the FASY-System based on human like description of each components of the face. The FASY System has the following features: (i) face queries using human-like description of the face, (ii) searching the required face in the database and (iii) generation of the requested face when they are not in the database. This paper is focused in this last feature. All the face components are stored in the database and are retrieved according to the user given queries and combined together to form the whole face which is then shown to the user. The Face Generation system works in 3 steps: (i) Searching phase to search the database for each single components of the face according to the user's query. (ii) Assembling phase to place all the components at proper places to generate the full face (iii) Tuning phase to adjust the intensities to cover up the stitching lines between different pair of face components so that the generated face looks natural.

A variety of face databases have already been described in literature and here we have worked with the databases found from AT & T Laboratories Cambridge [8]. The specifications for the database described in this paper are as follows:

- In all blank facial images, the ears should be visible.
- Sufficient variety of facial components and their properties should be stored in the database.
- The database should contain subject of different ethnicities, sample of both men and women, and sample from different age groups.
- Variation due to head tilt, shift, rotation and scaling should be minimized as much as possible.

## II. FACE COMPONENTS AND ASSOCIATED PARAMETERS

The human-like face description that FASY accepts has been determined by a psychological study. As a result of this study, we found 7 face components and a set of associated parameters to be used in our system. The face components are Face cutting (with ears and hairs), right eyebrow, right eye, left eyebrow, left eye, nose and lips. The different parameters and their values for each of the seven components of a face are as follows:

**A. Face cutting**
(I) Sex (Male, Female)
(II) Age (11-20, 21-30, 31-40, 41-50, 51-60, 61-70, Above 70)
(III) Shape (Oval, Round, Cant Say)
(IV) Jaw (Wide, Narrow, Normal, Cant Say)
(V) Hair Density (Highly Dense, Low Dense, Normal, Cant Say) Hair Color (Black, Brown, Cant Say)

**B. Eyebrows (Left and Right)**
(I) Length (Small, Large, Normal, Cant Say)
(II) Width (Small, Large, Normal, Cant Say)
(III) Shape (Flat, Round, Wavy, Artistic, Cant Say)
(IV) Hair (Highly Dense, Low Dense, Normal, Cant Say)

**C. Eyes (Left and Right)**
(I) Length (Small, Large, Normal, Cant Say)
(II) Width (Small, Large, Normal, Cant Say)

(III) Shape (Round, Elliptic, Cant Say)
(IV) Eye-boll color (Black, Brown, Green, Blue, Cant Say )
**D. Nose**
(I)  Sharpness (Sharp, Blunt, Normal, Cant Say)
(II) Nostrils (Small, Large, Normal, Cant Say)
(III) Length (Small, Large, Normal, Cant Say)
(IV) Width (Small, Large, Normal, Cant Say)
**E. Lip**
(I)  Length (Wide, Small, Normal, Cant Say)
(II)  Width (Thick, Thin, Normal, Cant Say)
(III) Surface (Smooth, Wrinkled, Cant Say)
(IV) Mouth (Open, Closed, Cant Say)
(V) Shape (Linear, Wavy, Cant Say)

### III. THE FASY SYSTEM

A human user makes a query of a face using a human-like face description. The FD module interprets the query and translates it into a face description using different Face Parameters. This face description is used by the FR module to search the face in the database. The system retrieves the existing faces based on the user description. Now the face may not exist in the database or the existing faces select the automatic generation of it. The FG module first finds the face components according to the user's requirements and shows them to the user. If may be rejected by the user. In the case when the desired face is not found in the database, the user can user is satisfied with the retrieved components then he/she can proceed for generating a new face with the selected face parameters. As a result of the generation process the generated face is presented to the user. If the user is satisfied with the generated face, the process stops here and the generated face is stored in the database. Otherwise, the user enters into an iterative process. The iterative generation of the faces is implemented in the PA module. Figure 1 shows the block diagram of FASY system.

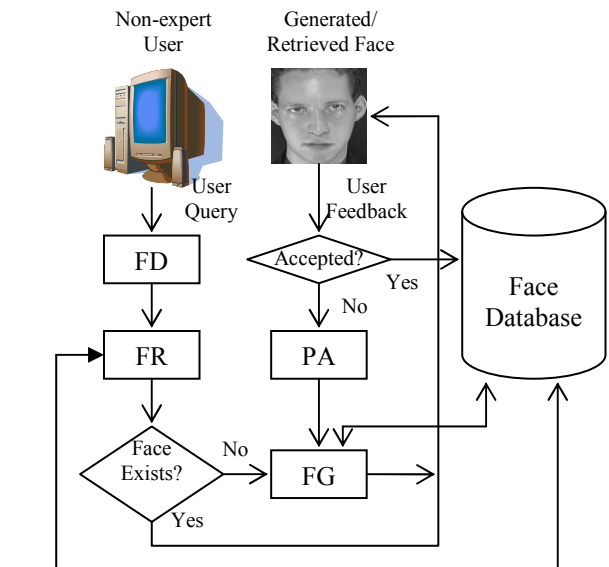

Non-expert User | Generated/Retrieved Face

**FD**: Face Descriptors      **FR**: Face Retrieval
**FG**: Face Generation       **PA**: Parameter Adjustment
Fig. 1: Block Diagram of FASY System

### IV. SEARCHING PHASE

The searching phase accepts the user description of the different components of the face and searches the database for each single components of the face like right), noses and lips. The parameters for the facial cuttings and components are stored in the database previously. For a set of given parameters of a component multiple images may be selected and user can choose one of them. Fig 2 shows the context diagram for Face Synthesis PA module.

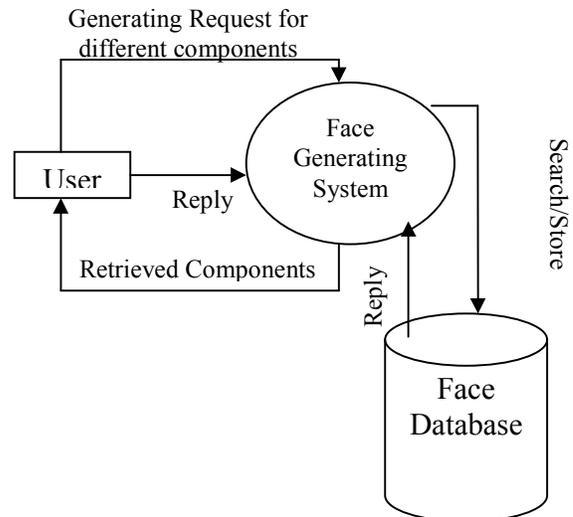

Fig 2: Context Diagram for Face Synthesis PA module

### V. ASSEMBLING PHASE

The retrieved components of the face selected by the user based on user's query are fed to the assembling phase to combine all the components to generate the full face. The assembling phase accepts all the components retrieved from the database and place those components on the appropriate position of the face. The arrangement of the facial features does not change very much from person to person; therefore we can predict the position of each facial feature on a face when the position of the ear is known. To get the proper position of the components first we find the right ear position and this is done by the algorithm 1.

*Algorithm 1*
1.   Algorithm Find_Position _Ear(FaceI)
2.   // FaceI is the intensity matrix of gray scale face.
3.   {
4.   Get the binary image of FaceI into a binary matrix *FaceBinary* with order m X n.
5.   // In binary image, 1 denotes white region and 0 denotes black region. So starting from first column find the first 1 value in FaceBinary matrix. Steps 6 to Step 14 do this.
6.   For j =1 to n
7.       For i = 1 to m
8.           If FaceBinary(i,j)=1
9.               Tx = i
10.              Ty = j

11.        Stop
12.     End if
13.     End for
14.   End for
15.  }// End of algorithm

Figure 3 shows a face and Figure 4 shows its binary image after intensity adjustment. Suppose (Tx, Ty) is the co-ordinates of the upper left corner of the right ear. The next task is to track the position of the other components. In general, eyes lie on the almost same x co-ordinate of the upper left corner of the ear. Here the size of each faces is 92X112 and for this size we have assumed some constant values for calculating the positions of the components on the blank face. Hence the co - ordinates of the components for a normal face are calculated as follows by the experimental results:

Tx = Tx;
Ty = Ty+10;
*Position of the right eyebrow* = (Tx−5, Ty)
*Position of the right eye* = (Tx, Ty+ width of the right eyebrow−width of the right eye)
*Position of the nose* = (Tx−2, Ty+ width of the right eyebrow)
*Position of the left eyebrow* = (Tx−5, y co - ordinate of nose + width of nose −5)
*Position of the left eye* = (Tx, y co - ordinate of left eyebrow)
*Position of the lip*= (x co - ordinate of nose + height of nose + 5, y co-ordinate of nose + width of nose/2 − width of lip/2)

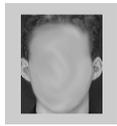     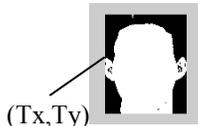

Fig. 3: Facial cutting        Fig. 4: Binary image

By the above procedure we find the normal positions of the face components. But for deformed shape of face these positions can vary and so this system has a provision to change the position of the components as required after tuning phase.

VI. TUNING PHASE

The retrieved components when overlapped on the basic face cutting, the background intensity of the face may not match with the facial intensity. To make the overall face clear, prominent and natural to the user, our system works in two steps:

1.  First get the binary image and from it find the actual image.
2.  Adjust the intensities so that the face looks natural.

Consider an image of an eye. Generally the shape of an eye is oval or round but dimension of the image of an eye is rectangular. So instead of copy the whole image we can store the intensities of oval region of the eye portions only. For this purpose binary image of the components are used. In binary image, value 1 denotes the white region and value 0 denotes the black region. In Figure 5, the black region contains the actual eye or actual eyebrow. So copy those intensities of the eye where the entries in the binary image of it contain 0.

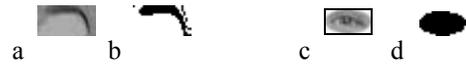

Fig. 5: (a) Original left eyebrow (b) Binary image of left eyebrow after intensity adjustment (c) Original left eye (d) Binary image of left eye after intensity adjustment

Now to match the intensities of the different components with face cutting intensities we have designed an algorithm which works with the neighborhood intensities. Here we have considered 3 X 3 neighborhood intensities.

| $FI_{(x-1, y-1)}$ | $FI_{(x-1, y)}$ | $FI_{(x-1, y+1)}$ |
|---|---|---|
| $FI_{(x, y-1)}$ | $FI_{(x, y)}$ | $FI_{(x, y+1)}$ |
| $FI_{(x+1, y-1)}$ | $FI_{(x+1, y)}$ | $FI_{(x+1, y+1)}$ |

Fig. 6: 3X3 intensity matrix of face cutting for the (x, y) position

| $CI_{(i-1, j-1)}$ | $CI_{(i-1, j)}$ | $CI_{(i-1, j+1)}$ |
|---|---|---|
| $CI_{(i, j-1)}$ | $CI_{(i, j)}$ | $CI_{(i, j+1)}$ |
| $CI_{(i+1, j-1)}$ | $CI_{(i+1, j)}$ | $CI_{(i+1, j+1)}$ |

Fig. 7: 3X3 intensity matrix of a face component for the (i, j) position

Now instead of copying the original intensities of a component into the face cutting, we calculate a new set of intensities for each component as follows:

Suppose the intensity value of facial cutting at the position (x,y) is $FI_{(x,y)}$ and the intensity value of a face component at the position (i,j) is $CI_{(i,j)}$ . $CI_{(i,j)}$ is to be copied at the (x,y) position on the face cutting. Now instead of replacing $FI_{(x,y)}$ by $CI_{(i,j)}$, we are calculating a new value of $CI_{(i,j)}$ for adjusting the intensity between face cutting and face components.

Summation of neighborhood intensities of face cutting at point (x,y) is

$F_1 = FI_{(x-1, y-1)} + FI_{(x-1, y)} + FI_{(x-1, y+1)} + FI_{(x, y-1)} + FI_{(x, y)} + FI_{(x, y+1)} + FI_{(x+1, y-1)} + FI_{(x+1, y)} + FI_{(x+1, y+1)}$

Summation of neighborhood intensities of a face component at point (i,j) is-

$C_1 = CI_{(i-1, j-1)} + CI_{(i-1, j)} + CI_{(i-1, j+1)} + CI_{(i, j-1)} + CI_{(i, j)} + CI_{(i, j+1)} + CI_{(i+1, j-1)} + CI_{(i+1, j)} + CI_{(i+1, j+1)}$

Intensity Factor IF = $F_1 / C_1$

Now calculate the new intensity of face cutting at the point (x,y) as -

$FI_{(x, y)} = (FI_{(x, y)} + 2* IF * CI_{(x, y)}) / (1 + 2 * IF)$

Algorithm 2 depicts the steps for tuning phase:

*Algorithm 2*
1.  Algorithm Place_Image(I, F)
2.  // I is intensity matrix of a face component with order m X n, $I_1$ is the binary matrix of the face component and F is the intensity matrix of facial cutting.
3.  {
4.  Get the binary image of I in the binary matrix $I_1$.
5.  For each row i of the matrix I (i=1 to m)
6.  For each column j of the matrix I (j=1 to n)
7.  If $I_1(i, j)$ is 0 then
8.  // Suppose intensity at (i, j) of the face component is to be copied at (x, y) position of the face cutting.
9.  FI = F(x-1, y-1) + F(x-1, y) + F(x-1, y+1) + F(x, y-1) + F(x, y) + F(x, y+1) + F(x+1, y-1)+F(x+1, y)+ F(x+1, y+1)
10. CI = I(i-1, j-1) + I(i-1, j) + I(i-1, j+1) + I(i, j-1) + I(i, j) + I(i, j+1)+ I(i+1, j-1) + I(i+1, j) + I(i+1, j+1)
11. IF = FI / CI
12. F(x, y) = (F(x, y) + 2 * IF * I(i, j)) / (1 + 2 * FI)
13. End if
14. End for
15. End for
16. } //End of Algorithm

Figure 8 shows the steps to construct a new face with the selected face cutting and face components by the user. The position of the face components are calculated using the algorithm 1 and the equations described in section 5.

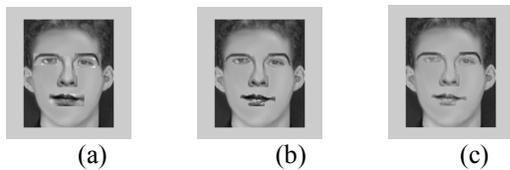

(a)          (b)          (c)

Fig. 8: Steps to generate a face

(a)Blind replacement of the components on the face (b) Replacing the actual image of each component (c) Final image after intensity adjustment of the components (Applying algorithm 2)

## VII. EXPERIMENTAL RESULTS

For testing the proposed method, we had about 200 male and female face images of different ages. From these images we obtained 100 gray scale facial cuttings, 50 gray scale left eyebrows, 50 gray scale left eyes, 50 gray scale right eyebrows, 50 gray scale right eyes, 30 gray scale noses and 30 gray scale lips. The different parameters for the facial cutting and components were set previously. In about 80% cases the FASY system generated the new faces successfully which fulfilled the user's query and looked natural. Suppose we want to generate a face which has the impressions described in Table 1:

Table 1: Face impressions of a person

| Components | Parameters | Values |
|---|---|---|
|  | Sex | Male |
|  | Age | 21-30 |

| Face Cutting | Shape | Oval |
|---|---|---|
|  | Jaw | Normal |
|  | Hair Density | Normal |
|  | Hair Color | Black |
| Right Eyebrow | Length | Large |
|  | Width | Normal |
|  | Shape | Elliptic |
|  | Hair | Highly Dense |
| Right Eye | Length | Normal |
|  | Width | Normal |
|  | Shape | Elliptic |
|  | Eye-boll Color | Black |
| Left Eyebrow | Length | Large |
|  | Width | Normal |
|  | Shape | Elliptic |
|  | Hair | Highly Dense |
| Left Eye | Length | Normal |
|  | Width | Normal |
|  | Shape | Elliptic |
|  | Eye-boll Color | Black |
| Nose | Sharpness | Normal |
|  | Nostrils | Normal |
|  | Length | Normal |
|  | Width | Normal |
| Lips | Length/Size | Normal |
|  | Width | Normal |
|  | Surface | Smooth |
|  | Mouth | Cant Say |
|  | Shape | Cant Say |

Figure 9 shows the snapshot of searching phase of FASY system for Table 1 and Figure 10 shows the snapshot of face generation by the FASY system using the selected face components.

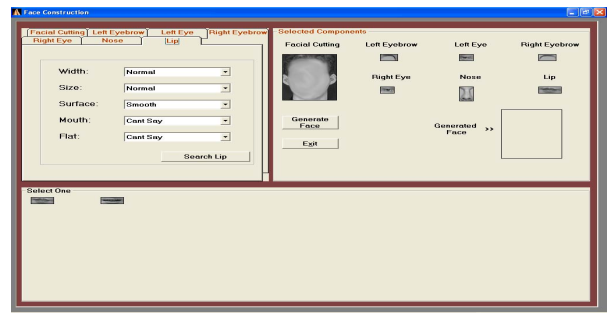

Fig 9: Snapshot of synthesis phase

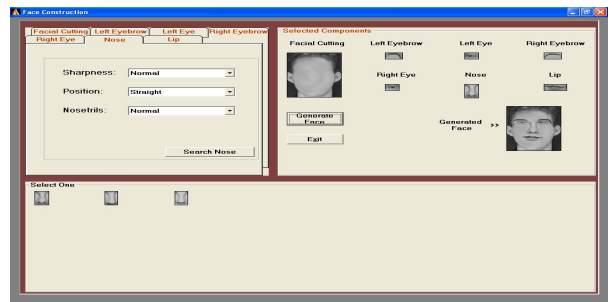

Fig 10: Snapshot of face generation

Table 2: Face impressions of another person

| Components | Parameters | Values |
|---|---|---|
| Face Cutting | Sex | Male |
| | Age | 31- 40 |
| | Shape | Oval |
| | Jaw | Normal |
| | Hair Density | Highly Dense |
| | Hair Color | Black |
| Right Eyebrow | Length | Small |
| | Width | Small |
| | Shape | Flat |
| | Hair | Low Dense |
| Right Eye | Length | Normal |
| | Width | Small |
| | Shape | Elliptic |
| | Eye-boll Color | Black |
| Left Eyebrow | Length | Small |
| | Width | Small |
| | Shape | Flat |
| | Hair | Low Dense |
| Left Eye | Length | Normal |
| | Width | Normal |
| | Shape | Elliptic |
| | Eye-boll Color | Black |
| Nose | Sharpness | Normal |
| | Nostrils | Normal |
| | Length | Normal |
| | Width | Normal |
| Lips | Length/Size | Normal |
| | Width | Normal |
| | Surface | Smooth |
| | Mouth | Closed |
| | Shape | Wavy |

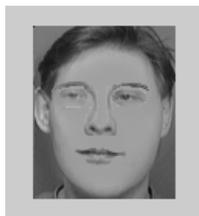

Fig. 11: Generated face according to Table 2

Table 2 depicts the face impressions of another person and the corresponding generated face is shown in Figure 11.

### VIII. CONCLUSION

Providing the facility of a user-friendly face generation system comes out to be a very complex task. The FASY system corresponds to an effort in this direction. The interactive retrieval process allows also an update of the face database. If the requested face can not be found in the database the user can choose to interactively generate it. The generation of the requested face when they are not found in the database is one of the main FASY characteristics. When the user interactively generates the face, the database can be expanded and the characteristics of the generated face that are supplied by the user are stored in the database as well. From this, even a small-at-the-beginning face database can grow from the very beginning on. This work can be especially useful for the criminal identification purpose where we have to construct a face of a criminal based on the description of eye-witness.

### ACKNOWLEDGMENT

Authors are thankful to the "Center for Microprocessor Application for Training Education and Research", "Project on Storage Retrieval and Understanding of Video for Multimedia" and Computer Science & Engineering Department, Jadavpur University, for providing infrastructural facilities during progress of the work. One of the authors, Mr. Santanu Halder, is thankful to Guru Nanak Institute of Technology for kindly permitting him to carry on the research work. Mr. Santanu Halder is also thankful to Infosys Technologies for their financial support through their Campus Connect program, Bhubaneswar.